\pdfoutput=1

\documentclass[11pt]{article}

\usepackage{acl}

\usepackage{times}
\usepackage{latexsym}

\usepackage[T1]{fontenc}

\usepackage[utf8]{inputenc}

\usepackage{microtype}

\usepackage{amsmath,amssymb,amsfonts}
\usepackage{cleveref}
\usepackage{comment}
\usepackage{graphicx}
\usepackage{epigraph}
\usepackage{notation}
\usepackage{pgf,tikz}
\usetikzlibrary{positioning}
\usepackage[most]{tcolorbox}
\title{Informativeness and Invariance:\\Two Perspectives on Spurious Correlations in Natural Language}

\author{Jacob Eisenstein \\
  Google Research \\
  \texttt{jeisenstein@google.com} \\}

\begin{document}
\maketitle
\begin{abstract}
Spurious correlations are a threat to the trustworthiness of natural language processing systems, motivating research into methods for identifying and eliminating them. However, addressing the problem of spurious correlations requires more clarity on what they are and how they arise in language data. \citet{gardner-etal-2021-competency} argue that the compositional nature of language implies that \emph{all} correlations between labels and individual ``input features'' are spurious. This paper analyzes this proposal in the context of a toy example, demonstrating three distinct conditions that can give rise to feature-label correlations in a simple PCFG. Linking the toy example to a structured causal model shows that (1) feature-label correlations can arise even when the label is invariant to interventions on the feature, and (2) feature-label correlations may be absent even when the label is sensitive to interventions on the feature. Because input features will be individually correlated with labels in all but very rare circumstances, domain knowledge must be applied to identify spurious correlations that pose genuine robustness threats.
\end{abstract}

\newcommand{\say}[1]{\textit{#1}}

\section{Introduction}
Spurious correlations are a growing source of concern in machine learning~\citep{geirhos2020shortcut} and related fields including natural language processing~\citep[][\textit{inter alia}]{gururangan2018annotation,mccoy2019right}. While the intuition is fairly clear --- spurious correlations are features that are useful in the training data but unreliable in general --- the notion is frequently referenced without a formal definition. \citet{gardner-etal-2021-competency} propose a definition in terms of conditional probabilities: a feature $X_i$ is spuriously correlated with the label $Y$ unless $P(Y\mid X_i)$ is uniform. The definition can be generalized from uniformity to independence ($X_i \indep Y$) without affecting the claims of the paper. They go on to argue that ``in a language understanding problem, \ldots \emph{all} simple correlations between input features and output labels are spurious'' (emphasis in the original). 
\begin{figure}
    \centering
    \begin{tikzpicture}[var/.style={draw,circle},
nonterm/.style={},
cfgedge/.style={densely dotted, thick, blue},
scmedge/.style={thin, ->}
]
\def\cfgheight{.3cm}
\def\scmheight{.3cm}
    \node (S) [nonterm] {$S$};
    \node (NP) [nonterm, below left=\cfgheight and 1cm of S.west]     {$\tilde{Z}_{\text{pizza}}$};
    \node (VP) [nonterm, below right=\cfgheight and 1cm of S.east]
    {$\tilde{Y}_{+}$};
    \node (X1) [below=\cfgheight of NP, ]{$\underbrace{\text{\say{The pizza\strut\vphantom{Ty}}}}_{X_1}$};
    \node (X2) [below left=\cfgheight and .33cm of VP, ]{$\underbrace{\text{\say{was not\strut\vphantom{Ty}}}}_{X_2}$};
    \node (X3) [below right=\cfgheight and .33cm of VP, ]{$\underbrace{\text{\say{too greasy\strut\vphantom{Ty}}}}_{X_3}$};
    
    \draw [cfgedge] (S) -| (NP);
    \draw [cfgedge] (S) -| (VP);
    \draw [cfgedge] (NP) -- (X1.north);
    \draw [cfgedge] (VP) -| (X2.north);
    \draw [cfgedge] (VP) -| (X3.north);
    
    \node (Z) [var, below=\scmheight of X1] {$Z$};
    \node (Y) [var] at (Z -| VP) {$Y$};
    \draw [scmedge] (X1.south) -- (Z.north);
    \draw [scmedge] (X2.south) -- (Y);
    \draw [scmedge] (X3.south) -- (Y);
    
\end{tikzpicture}
    \caption{An instance from the toy model. The upper part of the figure corresponds to $f_X$, the function that generates the text via a PCFG (see \cref{fig:causal-model}): nodes represent non-terminals in the grammar and edges represent context-free derivations. The lower part of the figure corresponds to the causal model of the sentiment $Y$ and target $Z$. Here nodes represent random variables and edges represent causal relationships.}
    \label{fig:cfg-and-scm}
\end{figure}
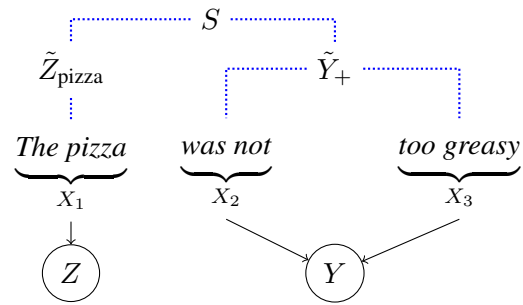
The property that individual input features should be independent of labels --- which I will call \emph{marginally uninformative input features (UIF)}\footnote{The features are \emph{marginally} uninformative because the criterion is the marginal distribution $P(Y | X_i) = \int P(Y, X_{\neg i} | X_i) d X_{\neg i}$. Features may be marginally uninformative while still giving information about the label when viewed in combination.} --- is treated 
as an assumption about the nature of language processing and also as a desideratum that datasets should satisfy: if the label can be predicted from input features alone, then the dataset is in some sense too easy.\footnote{To formalize the UIF assumption, it is necessary to clarify which features are ``input features'': bytes, phonemes, wordpieces, words, phrases, or sentences? 
The selection of input features is a property of the model and not the dataset, but the intuitive support for UIF seems stronger for features that are lower on the linguistic hierarchy. 
Because the arguments presented here don't depend on the specific definition of input features, I will follow \newcite{gardner-etal-2021-competency}, who informally identify input features with words. However, if one were to apply UIF for a practical purpose such as dataset curation, it would be important to explore this issue more thoroughly, particularly in regard to languages in which words are the sites of a significant amount of morphological composition and are therefore capable of carrying complex relational meanings. Conversely, multiword expressions can function analogously to single word features, so there is no reason in principle why only single-word features should be considered spurious~\citep{Schwartz:2022}.
}

The principle of UIF is based on the insight that linguistic context can modulate the semantics of any subspan of a text, using mechanisms such as syntactic negation or discourse markers. Furthermore, the frequency of negation and other forms of semantic inversion may vary across datasets and deployment settings. A predictor that relies on negation being rare (to pick one example) cannot be said to have truly achieved competence in the language processing task. Such a predictor may perform poorly in domains in which these high-level distributional properties shift.

An especially provocative assertion of \citeauthor{gardner-etal-2021-competency} is that all correlations between labels and individual input features have the same status. In the sentence \say{the pizza was amazing}, suppose that both \say{pizza} and \say{amazing} are correlated with positive sentiment because the reviewers like pizza. There are at least two intuitive differences between these two correlations. First, while one can easily imagine a benighted subpopulation of reviewers who do not like pizza, it is not so easy to imagine reviewers who think that the word ``amazing'' carries negative sentiment. Second, if we modify the subject (e.g., \say{the \textbf{movie} was amazing}), the label will usually be unaffected, but there are many perturbations to the adjective that flip the label (e.g., \say{the pizza was \textbf{greasy}}). This second intuition can be described using the framework of causality, which has generally treated spurious correlations as those that arise without a direct causal explanation~\citep{simon1954spurious}. Given a causal model of the data generating process, we can compute an \emph{interventional} distribution $P(Y \mid \text{do}(X_1 := x), X_2, X_3),$ which corresponds to the distribution over $Y$ in a data generating process in which the variable $X_1$ is surgically set to the value  $x$~\citep{pearl1995causal,peters2017elements,feder2021causal}.\footnote{Space does not permit a discussion of the distinction between interventions and counterfactuals~\citep[see][]{pearl2009causality}.} When such interventions do not affect $Y$ for any given example, we say that $Y$ and $X_1$ are \emph{counterfactually invariant}~\citep{veitch2021counterfactual}. Violations of UIF are particularly troubling when they are accompanied by counterfactual invariance, because non-causal correlations often do not transfer to other domains~\citep{scholkopf2012causal,buhlmann2020invariance}.

This paper uses a toy example to relate the UIF property to (1) the production probabilities in probabilistic context-free grammars (PCFGs), and (2) counterfactual invariance in structured causal models. The connection to PCFGs provides additional motivation for the UIF criterion from the perspective of domain generalization, while clarifying the scenarios that can give rise to violations of UIF, which \citeauthor{gardner-etal-2021-competency} attribute too narrowly to ``bias and priming effects'' in annotators. The connection to counterfactual invariance highlights the ways in which these concepts do and do not align. Efforts to remove artifacts from the training and evaluation of NLP systems will be most productive when focused at the intersection of these two views of spurious correlations: violations of UIF for input features to which the label is counterfactually invariant according to a plausible causal model.

\section{Toy Example}
\label{sec:example}
Consider a simplified targeted sentiment analysis task~\citep{mitchell2013open}, in which the sentiment is $Y$, the target is $Z$, and the sentences are all of the form $(X_1, X_2, X_3)$, with $X_1$ specifying a target noun phrase, $X_2$ a copula-like expression, and $X_3$ a predicative adjectival phrase. For example, ${Y = \textsc{Pos}}, {Z = \textsc{Pizza}}, {X_1 = \text{\say{the pizza}}}, X_2 = \text{\say{turned out to be}}, {X_3 = \text{\say{crispy and delicious}}}.$
We will treat this data as generated from the causal model shown in \cref{fig:causal-model}.
This causal model can be summarized by two assertions: (1)
the target $Z$ is a direct effect of only the span $X_1$; (2) the sentiment label $Y$ is a direct effect of only the spans $X_2$ and $X_3$.  
The function $f_X$ can represent any generative model of text: an n-gram model, a grammar-based formalism, a deep autoregressive network, etc.

\begin{figure}[]
    \centering
\begin{tcolorbox}[ams align,colback=white]
U := & N_U \\
(X_1, X_2, X_3) := & f_X(U, N_X)\\
Z := & f_Z(X_1, N_Z)\\
Y := & f_Y(X_2, X_3, N_Y).
\end{tcolorbox}
    \caption{Causal model for the toy example shown in \cref{fig:cfg-and-scm}. $N_U, N_X, N_Y, N_Z$ indicate independent noise variables, and $f_X, f_Y, f_Z$ indicate deterministic functions that map from causes to effects~\citep[for more details on the notation, see][]{peters2017elements}.}
    \label{fig:causal-model}
\end{figure}

\paragraph{Aside on the direction of causation.} We treat the text as the cause of the labels, rather than the converse. This distinction is somewhat vexed~\citep{scholkopf2012causal,jin-etal-2021-causal}. In some cases the direction of causation is clear from the task (e.g., table-to-text generation, summarization, and translation), but often the problem could be framed in either direction: perhaps the writer had the label in mind when producing the text, and thus the text is an effect of the label; or perhaps it is better to think of the annotator, who must read the text to arrive at the label, regardless of the writer's intentions.
When the labels cause the text, the notion of counterfactual invariance can be restated in terms of the invariance of text features to perturbations on labels, e.g. $P(X_1 \mid \text{do}(Y := y), Z)$.
As the toy example is meant to serve only an expository purpose, we leave elaboration of the relationship of UIF to such models for future work.

\subsection{Counterfactual invariance $\nRightarrow$ UIF}
The causal model implies several counterfactual invariance properties: intervention on $X_1$ will not affect $Y$, nor will intervention on $X_2$ or $X_3$ affect $Z$. This is because $X_1$ blocks the influence of $X_2$ and $X_3$ on $Z$, and vice versa for $Y$. 
Conversely, $(X_3, Y)$ are not counterfactually invariant in general because $X_3$ is an ancestor of $Y$ in the causal graph, and similarly for $(X_2, Y)$ and $(X_1, Z)$. 

Counterfactual invariance does not imply that the associated input features are marginally uninformative of the label.
Consider a classical spurious correlation in which pizza tends to receive positive sentiment and sushi receives negative sentiment. This correlation is produced when $f_X$ encodes a PCFG with the top-level production:
\begin{equation}
\begin{split}
S \to \quad & \tilde{Z}_{\text{pizza}} \; \tilde{Y}_{+} \qquad (1 + \alpha) /4\\
& \tilde{Z}_{\text{sushi}} \; \tilde{Y}_{-} \qquad (1 + \alpha) / 4\\
& \tilde{Z}_{\text{pizza}} \; \tilde{Y}_{-} \qquad (1 - \alpha) / 4\\
& \tilde{Z}_{\text{sushi}} \; \tilde{Y}_{+} \qquad (1 - \alpha) / 4,
\label{eq:s-production}
\end{split}
\end{equation}
with the right column indicating the probability of each rule expansion and $\alpha \in [-1, 1]$.\footnote{The stochasticity of the grammar is encoded in the deterministic function $f_X$ through the noise variable $N_X$. Let $N_X \sim \text{Uniform}(0, 1)$, and choose the first rule expansion of $S$ when $N_X < (1 + \alpha) / 4$, the second rule expansion when $(1 + \alpha) / 4 \leq N_X < (1 + \alpha) / 2$, and so on.} The nonterminal symbols $\tilde{Z}_{\text{pizza}}, \tilde{Z}_{\text{sushi}}, \tilde{Y}_{+}, \tilde{Y}_{-}$ are intentionally chosen to correspond to the labels $Z$ and $Y$.
Subsequent rules in the grammar can then be designed to ensure that $\tilde{Z}_{\text{pizza}}$ usually produces values of $X_1$ that make $Z = \textsc{Pizza}$ likely, and analogously for the other non-terminals and associated labels. The unification of PCFGs and structured causal models is shown in \cref{fig:cfg-and-scm}.

When $\alpha \neq 0$, there may be an association between $X_1$ and $(X_2, X_3)$.
As a result, there exist pairs of values $(x_1, x'_1)$ such that,
\begin{equation}
\begin{split}
    P & (Y | X_1 = x_1) \\
    & = \sum_{X_2, X_3} P(Y \mid X_2, X_3) P(X_2, X_3 \mid X_1 = x_1)\\
& \neq \sum_{X_2, X_3} P(Y \mid X_2, X_3) P(X_2, X_3 \mid X_1 = x'_1) \\
& = P(Y | X_1 = x'_1),
\end{split}
\end{equation}
creating a violation of UIF. The same argument can be applied to $P(Z \mid X_2)$ and $P(Z \mid X_3)$. UIF is also violated in $P(Z \mid X_1)$, $P(Y \mid X_2)$, and $P(Y \mid X_3)$, but for a different reason: these distributions are conditioned on the direct causal parents of the labels in $f_Y$ and $f_Z$. Manipulation of the data distribution to ensure that $\alpha = 0$ (deconfounding $\tilde{Y}$ and $\tilde{Z}$) can remove only the violations of UIF induced by $f_X$, but not those induced by the direct causal relationships encoded in $f_Y$ and $f_Z$: for example, if $\Pr(X_3 = \say{delicious} | \tilde{Y}_+) > \Pr(X_3 = \say{delicious} | \tilde{Y}_{-})$ then the feature \say{delicious} will be associated with positive sentiment regardless of the rule probabilities in \cref{eq:s-production}.

\paragraph{Discussion.}
The example shows how violations to UIF can emerge via confounding, creating classical spurious correlations in the sense of \citet{simon1954spurious}: informativeness despite counterfactual invariance. Such correlations are unlikely to be robust because it is not difficult to imagine a domain in which the sign of $\alpha$ changes, impairing the performance of predictors that have learned the spurious correlation. In contrast, feature-label correlations that arise directly from the causal model, such as $(Z, X_1)$, are only damaging under more extreme forms of concept shift, in which the meanings of the features themselves change.

\paragraph{Aside on causality and robustness.} The distinct interpretations of spuriousness as (1) non-causal and (2) non-robust are noted by \citet{Schwartz:2022} in concurrent work. However, these interpretations can be reconciled by the argument that non-causal features are inherently unlikely to be robust, which is sometimes formalized as the \emph{principle of sparse mechanism shift}~\citep{scholkopf2021toward}. The principle states that complex causal systems are usually composed of smaller independent parts, with domain shifts affecting only a few components of the system at a time. A related principle arises in the context of natural language: distributional frequencies are more likely to change across domains, while categorical facts about language are generally stable. \citet{biber1991variation}, for example, makes this argument explicitly in the analysis of register. In our model, the implication is that the probabilistic rule expansions in $f_X$ are more likely to change than the basic properties of the lexicon, which govern which terminal symbols can be emitted by each non-terminal.

\subsection{UIF $\nRightarrow$ Counterfactual Invariance}
Violations of counterfactual invariance can occur even when UIF is satisfied. To show this, we supply two more productions for the grammar:
\begin{align}
\begin{split}
    \tilde{Y}_{+} \to \quad & \textsc{Cop}_{+} \; \textsc{AdjP}_{+} \qquad \beta_{+}\\
     & \textsc{Cop}_{-} \; \textsc{AdjP}_{-} \qquad 1 - \beta_{+}\\
     \end{split}
     \label{eq:y-pos-production}\\
    \begin{split}
    \tilde{Y}_{-} \to \quad & \textsc{Cop}_{+} \; \textsc{AdjP}_{-} \qquad \beta_{-}\\
     & \textsc{Cop}_{-} \;  \textsc{AdjP}_{+} \qquad 1 - \beta_{-}\\
     \end{split}
     \label{eq:y-neg-production}
     \end{align}
Here the non-terminal $\textsc{Cop}_{+}$ produces a ``positive'' copula in $X_2$ (\say{is}, \say{was}, \say{is universally agreed to be}), $\textsc{Cop}_{-}$ produces a negated copula in $X_2$ (\say{isn't}, \say{wasn't}, \say{was the furthest possible thing from}), $\textsc{AdjP}_{+}$ produces positive-sentiment adjectival phrases in $X_3$ (\say{great}, \say{delicious}), and $\textsc{AdjP}_{-}$ produces negative-sentiment adjectival phrases in $X_3$ (\say{disappointing}, \say{totally unappetizing}). There are two special cases of interest: 
\begin{itemize}
\setlength\itemsep{0pt}
    \item When $\beta_+ = \beta_-$, the probability of using a negated copula is independent of $Y$, so $X_2$ satisfies UIF with regard to $Y$, while $X_3$ generally does not. 
    \item When $\beta_{+} = 1 - \beta_{-}$, the use of negation is balanced to make the distribution over sentiment terms independent of $Y$, so $X_3$ satisfies UIF with $Y$, while $X_2$ generally does not. 
    \end{itemize}
Combining these cases, both $X_2$ and $X_3$ satisfy UIF with $Y$ when $\beta_{+} = \beta_{-} = \frac{1}{2}$, meaning that negated and non-negated copula are equally likely and are independent of $Y$. 

\paragraph{Discussion.}
UIF is violated not only by confounding, as discussed in the previous section, but also in mild settings that do not meet any reasonable definition of bias: unless $\beta_+ = \beta_- = 1/2$ then at least one of $X_2$ and $X_3$ is marginally informative of $Y$.
Furthermore, UIF has no impact on the counterfactual invariance of $X_2$ and $X_3$ on $Y$. Neither is counterfactually invariant even when the generative model is parametrized to make UIF hold for all input features~\citep[see also][page 185]{pearl2009causality}. This is because the overall sentiment can be directly affected by adding or removing negation and by flipping the polarity of the sentiment-carrying adjective. 

\section{Conclusions}
In the toy example, violations of UIF arise from three distinct phenomena: confounding between the sentiment and the target ($\alpha \neq 0$, leading to $X_1 \notindep Y$); confounding between the sentiment and the use of negation ($\beta_+ \neq \beta_-$, leading to $X_2 \notindep Y$); and
lack of a perfect balance in the probability of negation between positive- and negative-sentiment examples ($\beta_+ \neq 1 - \beta_{-}$, leading to $X_3 \notindep Y$). The conditions required to satisfy UIF are thus progressively less plausible as we move from $X_1$ to $X_3$, and full UIF is achieved only in the perfectly balanced case of $\alpha=0, \beta_{+} = \beta_{-} = \frac{1}{2}.$ The number of such constraints will increase with the size of the grammar, making UIF vanishingly rare in more general settings. 
This conclusion follows from the PCFG analysis and is derived without reference to causality.

The toy example also demonstrates the disconnect between the UIF view of spurious correlations and the causal view: counterfactual invariance does not imply UIF because $X_1$ can be marginally informative of $Y$ even when $X_1$ and $Y$ are counterfactually invariant (these are the artifacts that we want to remove); UIF does not imply counterfactual invariance because both $X_2$ and $X_3$ can be uninformative of $Y$ even when $Y$ is sensitive to interventions on both features. From a theoretical perspective, it is unsurprising that these two views diverge, because UIF is a purely observational criterion while counterfactual invariance requires an explicit causal model. Indeed, this relationship is discussed in depth by \citet[][\S 6.3]{pearl2009causality}, albeit outside the context of language. The two perspectives can be seen as complementary, in that violation of UIF is a necessary but insufficient condition for a spurious correlation in the causal sense.

Moving beyond toy examples, it is unlikely that we can construct fully-specified causal models of language that supply useful invariances while handling every possible fluent utterance. How then can we use causal insights to design better benchmarks and more robust language understanding systems?
In some cases it is possible to elaborate partial causal models of a task, with associated invariance properties: for example, the sentiment of a movie review should be invariant to (though not independent of) the identities of the actors in the movie. Several existing approaches can be viewed as instantiations of partial causal models: for example, data augmentation, causally-motivated regularizers, stress tests, and ``worst-subgroup'' performance metrics (and associated robust optimizers) can be seen as enforcing or testing task-specific invariance properties that provide robustness against known distributional shifts~\citep[e.g.,][]{lu2020gender,ribeiro-etal-2020-beyond,kaushik2021explaining,koh2021wilds,veitch2021counterfactual}. Such approaches generally require domain knowledge about the linguistic and causal properties of the task at hand --- or to put it more positively, they make it possible for such domain knowledge to be brought to bear. Indeed, the central argument of this paper is that no meaningful definition of spuriousness or robustness can be obtained without such domain knowledge.

A final observation, pertaining to both UIF and counterfactual invariance, is the parallel treatment of $X_2$ (the copula) and $X_3$ (the adjectival phrase). From a lexical semantic perspective, only $X_3$ is directly associated with sentiment, while $X_2$ plays a functional role by potentially reversing $X_3$. It may therefore seem undesirable to learn a correlation between $X_2$ and $Y$, and preferable to attach that relationship exclusively to $X_3$. Indeed, one of the main catalysts of interest in spurious correlations in natural language processing was the observation that the presence of syntactic negation is a strong predictor of contradiction label in the natural language inference task, which should require reasoning about pairs of sentences~\citep{gururangan2018annotation,poliak2018hypothesis}. Yet neither UIF nor counterfactual invariance is capable of making any distinction between $X_2$ and $X_3$ in this model.  While it is possible to enforce uninformativeness on $X_2$ heuristically, e.g. by sampling or augmenting the data to ensure $\beta_+ = \beta_-$,
those same heuristics could be applied to enforce uninformativeness on $X_3$ by making $\beta_+ = 1 -\beta_-$. Singling out $X_2$ requires additional justification. Such a principle might be found in the multitask setting, in which we prefer feature-label informativeness to be sparse, with each feature directly informing only a few labels.

\paragraph{Acknowledgments.} Thanks to Alex D'Amour, Amir Feder, Katja Filippova, Matt Gardner, Tal Linzen, Ellie Pavlick, Noah Smith, Ian Tenney, Kristina Toutanova, and Victor Veitch for feedback on this work.

\bibliography{custom}
\bibliographystyle{acl_natbib}

\end{document}